\documentclass[nouppercase]{ifmbe} 

\makeatletter
\renewcommand{\author}[3][]{
      \stepcounter{ifmbe@authors}
      \expandafter\def\csname ifmbe@author\alph{ifmbe@authors}\endcsname
      {#2$^{\expandafter\the\csname ifmbe@affiliationcounter#3\endcsname
        \if\relax\detokenize{#1}\relax\else,#1\fi}$}
}
\makeatother

\title{An automatic deep learning approach for coronary artery calcium segmentation}

\affiliation{Department of Information Engineering, University of Pisa, Pisa, Italy }{FIRSTAFF}
\affiliation{U.O.C. Bioengineering, Fondazione Gabriele Monasterio, Massa, Italy }{SECONDAFF}
\affiliation{Imaging department, Fondazione Gabriele Monasterio, Massa, Italy }{THIRDAFF}

\author[2]{G. Santini}{FIRSTAFF}
\author{D. Della Latta}{SECONDAFF}
\author{N. Martini}{SECONDAFF}
\author{G. Valvano}{FIRSTAFF}
\author{A. Gori}{SECONDAFF}
\author{A. Ripoli}{SECONDAFF}
\author{C.L. Susini}{THIRDAFF}
\author[2]{L. Landini}{FIRSTAFF}
\author{D. Chiappino}{THIRDAFF}

\begin{document}
\nocite{*}

\maketitle

\begin{abstract}
Coronary artery calcium (CAC) is a significant 
marker of atherosclerosis and cardiovascular events. In this 
work we present a system for the automatic quantification of 
calcium score in ECG-triggered non-contrast enhanced cardiac 
computed tomography (CT) images. The proposed system uses 
a supervised deep learning algorithm, i.e. convolutional neural 
network (CNN) for the segmentation and classification of candidate 
lesions as coronary or not, previously extracted in the 
region of the heart using a cardiac atlas. We trained our network 
with 45 CT volumes; 18 volumes were used to validate the model 
and 56 to test it. Individual lesions were detected with  a sensitivity 
of 91.24\%, a specificity of 95.37\% and a positive predicted 
value (PPV) of 90.5\%; comparing calcium score obtained by the 
system and calcium score manually evaluated by an expert operator, 
a Pearson coefficient of 0.983 was obtained. A high agreement 
(Cohen's $\kappa$ = 0.879) between manual and automatic risk 
prediction was also observed. These results demonstrated that 
convolutional neural networks can be effectively applied for the 
automatic segmentation and classification of coronary calcifications.

\end{abstract}


\section{Introduction}

Cardiovascular disease (CVD), especially coronary heart 
disease (CHD), is one of the leading cause of death in the western 
world \cite{Palmieri2013}. It has been demonstrated that including 
coronary artery calcium score in traditional risk factors, 
results in a significant improvement in the classification 
of risk for the prediction of CHD events \cite{Polonsky2010}.\\
Usually, in clinical practice coronary calcium is manually detected 
by expert operators on ECG-triggered non-contrast enhanced 
CT images, i.e. without the use of contrast medium, 
with the aid of semi-automatic software. Typically, the identified 
lesions are characterized by intensity values above a standard 
threshold of 130 HU and belonging to coronary artery's structures. 
Subsequently, on the basis of all detected objects, 
the Agatston score \cite{Agatston1990} is computed with the aim of defining 
the corresponding cardiac risk. Even though this manual 
approach represents the clinical standard, it remains a time-consuming  
and operator dependent task.

In recent years a remarkable progress has been made in automatic 
image recognition, primarily due to a class of deep 
learning algorithms namely convolutional neural networks \cite{LeCun1998} 
(CNNs or ConvNets), which have shown great performance 
in computer vision tasks, such as object detection and image 
analysis \cite{Krizhevsky2012}. The main advantage of a CNN is related 
to its deep architecture, characterized by a series of layers 
able to extract from raw data a set of discriminative features, 
not designed by human engineers \cite{LeCun2015}, 
in order to operate classifications tasks or image captions.

In this work we exploited CNN to develop an automatic 
calcium scoring system,able to identify true coronary calcifications 
and discard other lesions, on ECG-gated basal CT acquisitions.
The performance of the CNN-based  automatic method for 
coronary calcium scoring was evaluated and compared with 
reference standard calcium score values obtained with 
manual annotations.

\section{Materials and Methods}

\begin{figure*}[ht]
      \centering
          \includegraphics[width=0.95\textwidth]{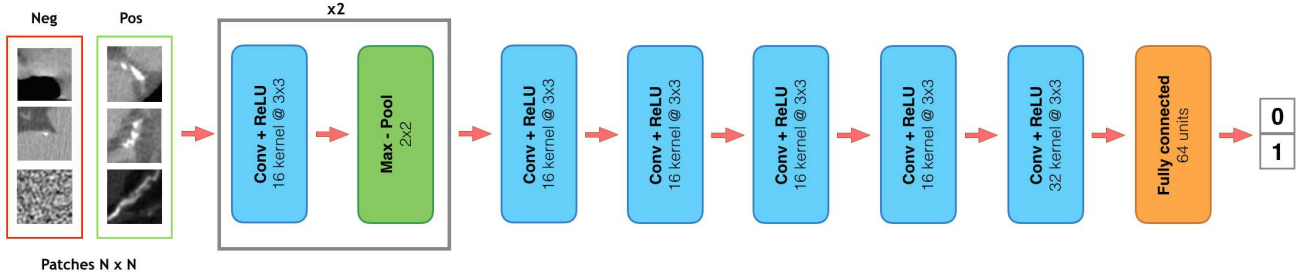}
      \caption{Block diagram of the convolutional network.}
      \label{figure1}
\end{figure*}

\subsection{Data description}

The acquisition protocol for calcium scoring involved 
ECG-triggered non-contrast enhanced CT images with a slice 
thickness of 3.0 mm, acquired with a tube voltage of 120 kVp 
and with different in-plane fields of view according 
to the patient heart's dimensions. 

This study included 152 exams coming from a screening 
study (MHELP) \cite{Pastormerlo2016} and randomly divided as follow. The 
first 33 CT scans were used to define a cardiac atlas for the extraction 
of the region of the heart. The remaining volumes were 
intended for the operations of the neural network and were 
split into a set of 45 exams to train the CNN from scratch, a 
set of 18 exams for the optimization of the network's hyperparameters 
and finally a set of 56 exams were used to evaluate 
the predictions performed by the system.

As reference, manual annotations of coronary calcifications 
were provided by an expert operator, with the Agatston 
score calculated for each patient according to $\sum\limits_{i}  a_i  \cdot \rho(a_i) \cdot \frac{\Delta z}{3.0}$, 
where $a_i$ is the area ($mm^2$) of the i-th coronary lesion, $\rho$ is 
a density factor depending on the largest intensity pixel in 
that plaque (130-199 HU: $\rho$=1, 200-299: $\rho$=2, 300-399: $\rho$=3,  
$\geq$400 HU: $\rho$=4), and $\frac{\Delta z}{3.0}$ represents a slice correction factor 
linked to the slice thickness of the acquisition ($mm$).\\
The final score was then used to categorize the patient's risk 
according to five classes:  0: no evidence (class A),  1-10: minimal 
(class B), 11-100: mild (class C), 101-400: moderate 
(class D), $>$400: severe (class E).

\subsection{Image processing}

A preliminary processing was performed on all the 
CT scans with the aim of reducing the computational load the 
CNN would had to deal with. This stage involved the definition 
of a cardiac atlas to be registered with the patients 
volumes, in order to restrict the identification of all the possible 
candidate lesions, only to the region of the heart.

For the creation of the cardiac atlas we based our approach 
on the image registration proposed by \cite{Avants2008}, a method 
which mainly relies on a multi-resolution strategy and performs 
iteratively alignment of volumes by means of affine and diffeomorfic 
registrations. To obtain a cardiac region of interest 
(ROI) for each patient, a single manual segmentation on the 
final atlas was performed to identify the heart. The resulted 
structure was then registered to all the CT volumes to remove 
everything outside the cardiac ROI. At this point to define all the 
possible candidate coronary lesions, a threshold of 130 HU 
was applied on the segmented CT scan. The detected candidate 
lesions were so reduced to true coronary calcifications, aortic 
and valves lesions, noise and possible image artifacts.

\begin{figure*}[ht]
      \centering
          \includegraphics[width=0.93\textwidth, height=5cm]{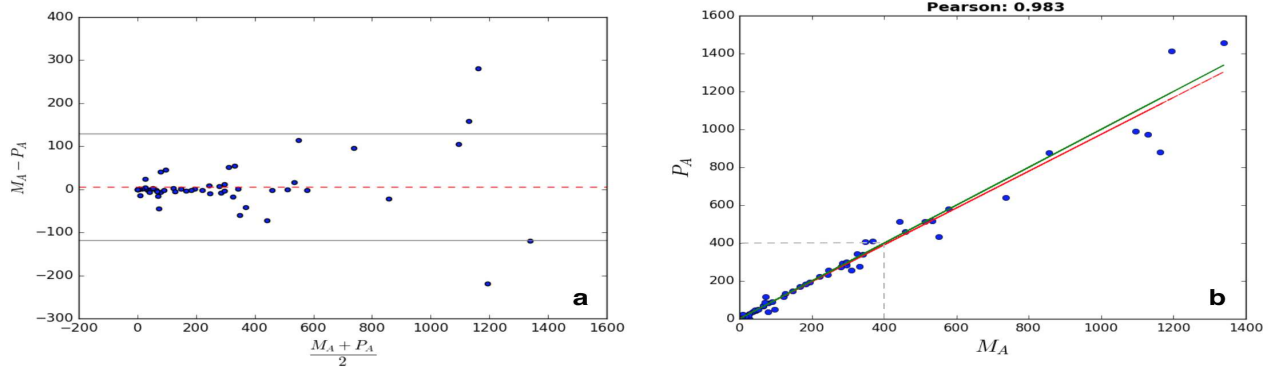}
      \caption{Agreement between Manual Agatston ($M_A$) and the Predicted Agatston ($P_A$). a) Bland-Altman plot with a 95\% confidence interval, b) the regression line (red) and the Pearson coefficient; dotted line separates the last risk class from the first four.}
      \label{figure2}
\end{figure*}

\subsection{Patch extraction and CNN design}

A specific classifier based on convolutional neural networks 
was implemented to correctly discriminate all candidate lesions 
extracted.

As proposed in \cite{Prasoon2013}\cite{Long2015} the designed ConvNet was fed with 
bidimensional patches, taken only from the axial projections 
of each scan and centered at the candidate pixel we want to 
classify as belonging or not to a coronary calcification. 
Since the original CT acquisitions were characterized by a wide 
range of in-plane resolutions (from 0.33 mm to 0.51 mm) all 
the volumes were resampled to achieve a 0.5 mm resolution 
in x-y plane. Nearest-Neighbor interpolation was used 
to avoid the creation of new intensity values which could 
affect the density factor $\rho$ in the Agatston score formula. Pixel 
spacing along z-axis was left to the routinely used 3 mm. 
After that, a series of patch sized at 25.5 x 25.5 mm (51 x 51 pixels) 
was cropped around each pixel belonging to that candiate 
lesions detected by the previous thresholding operation.\\ 
Due to the high imbalance between positive and negative samples, 
the ConvNet was trained with balanced batches. 
In addition, for each batch, half of the negative cases 
corresponded to patches centered at aortic calcifications, which 
represented the main source of false positive considering the 
heart as region of interest. In both training and test phase all 
the patches were normalized by mean value and standard 
deviation to allow a more robust training and a better prediction.

The architecture of the convolutional network used in this 
study was inspired by \cite{Wolterink2016} and implemented using Theano 
framework \cite{Team2016}. The feed-forward neural network consisted 
of seven convolutional layers (Fig.~\ref{figure1}). Each layer was 
characterized by 16 kernels of size 3x3, with the exception of the 
last one in which 32 kernels were applied. All the convolutions 
were valid mode, which means we retained the middle 
part of the full result of the convolution, thus avoiding 
from facing boundary situations. The inclusion of 2x2 max-pooling 
after the first two convolutional layers guaranteed the 
possibility to reduce the number of the network parameters 
and to introduce spatial invariance. Finally a single fully connected 
layer with 64 units was connected to the output of the 
ConvNet through a softmax function. \\
The softmax classifier produced a normalized probability (pCAC), 
which was thresholded at 0.5 in order to define the 
candidate pixel as either CAC on non-CAC.
ReLU activation function \cite{Glorot2011} and uniform He initialization for weights~\cite{He2015} 
were used throughout. The default value of the learning rate 
was set to 0.001, while the weights update was computed 
using Stochastic Gradient Descent (SGD) with Adagrad optimizer.
To prevent network overfitting, two regularization strategies were employed: 
a Dropout method with a probability of 0.5 was applied on the fully connected layer 
and an early stopping strategy was set during the validation phase. 

For a better management and retrieving of the data, the 
extracted patches were stored in a database together with the 
label of the central pixel, assigned by an expert operator, 
and its coordinates.

\subsection{Evaluation metrics}

Sensitivity or true positive rate, was computed on a single 
scan as the number of true positives divided by the total 
number of positives. The result was then averaged on the entire 
test set. To evaluate the number of false positives, which 
represented the main source of error in the network performance, 
positive predictive value (PPV) and specificity were employed.

In addition to the above statistics, used to evaluate the ability 
of the proposed method to correctly classify individual 
coronary lesions, Agatston score was computed by the system 
to quantify the amount of coronary artery calcium. 
The agreement in determining the Agatston-based class risk, 
between the automatic prediction and the expert operator's evaluation, 
was quantified by a linearly weighted Cohen's $\kappa$. On 
the other hand, to compare the predicted Agatston score with 
the manually defined ones, we computed the Pearson coefficient 
and we generated a Bland-Altman plot with a 95\% 
confidence interval (Fig.~\ref{figure2}).

\section{Results}

The presented method yielded a good sensitivity (91.24\%) 
in detecting coronary lesions per scan, and reached high 
specificity (95.37\%) with a satisfying PPV (90.50\%) in 
discriminating false coronary lesions from real ones.

The Agatston-based risk assessment of  56 patients was 
calculated and compared with the manual annotations 
provided by an expert operator, considered as the ground truth 
reference. Our automatic method achieved 91.1\% risk categorization 
accuracy with a linearly weighted Cohen's $\kappa$  value 
of 0.879 and a Pearson coefficient of 0.983 (Fig.~\ref{figure2}).

The system trained on a single NVIDIA Tesla 2070 GPU 
for about 1200 epochs with a time step of 3000 s/epoch. On 
the same GPU it reached fast performance in prediction task, 
with an average classification time of 0.78 $\pm$ 0.12 s per CT volume.

\section{Discussion}

The proposed method, exploiting the predictive power of 
CNNs, provides an automatic system for the quantification of calcium score.
\begin{figure}[ht]
      \centering
          \includegraphics[width=0.48\textwidth, ]{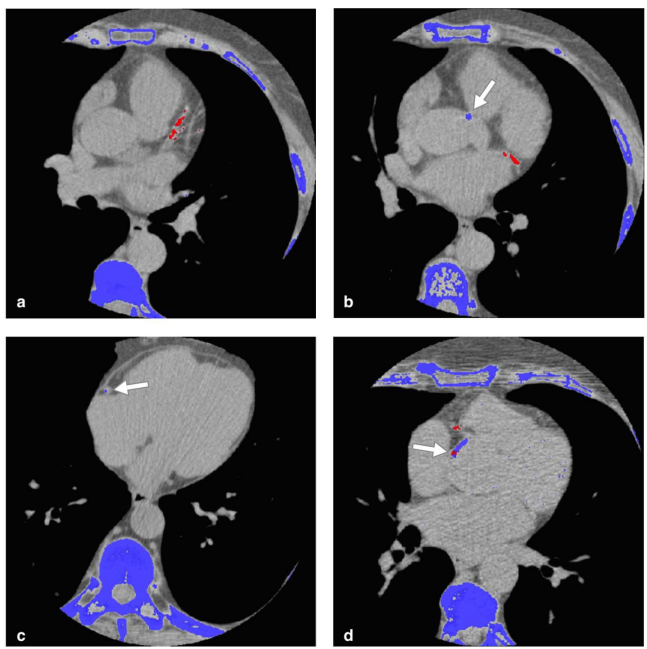}
      \caption{Examples of classification, where are reported pixels classified as positive by the CNN (red) and pixels above 130 HU (blue). a) Correct coronary predictions. b) Correctly discarded lesion in the ascending aorta. c) Missed calcification. d) False positive on the ascending aorta.}
      \label{figure3}
\end{figure}
As we observe in Fig.~\ref{figure2} the network had a 
better accuracy for the first four classes compared to the last one. 
However, a lower prediction sensitivity did not imply a 
wrong risk classification for the cases belonging to the last class. 
Within this context the patch undersampling to 0.5 mm 
of in-plane resolution, adopted in this work, did not affect 
negatively the Agatston-based risk assessment.

The analysis focused on the cardiac region, instead of 
processing the entire image, helped to reduce false positive errors, 
typically generated on the ribs and in some cases on the 
descending aorta. Moreover, the definition of a limited region 
of interest guaranteed a faster prediction on a single volume, 
thanks to the reduced number of pixels to be classified.

The classification task was demanded to a CNN with a
slightly different architecture from \cite{Wolterink2016}. In our CNN the presence 
of two max pooling prevented undersegmentation errors in some coronary lesions.
Interestingly, the network was able to distinguish between an ascending 
aorta lesion and a proximal coronary ones (Fig.~\ref{figure3}). 
This is probably due to the proposed strategy in the definition of 
the training set, where half of the negative cases had aortic lesions . 
As a result of this specific choice of negative candidates, 
instead of a completely random selection, the network 
learned to better recognize and discharge most of the false 
positives coming from the ascending aortic.

Despite the proposed CNN was trained with a small sample, 
the results obtained highlight that this method can effectively 
be used for the automatic segmentation and classification of 
coronary calcifications and can potentially achieve improved 
results in larger case studies.

\section{Conclusion}

In this study an automatic system based on a convolutional 
neural networks was successfully applied to the automatic 
segmentation and classification of coronary calcifications 
from ECG-gated non contrast enhanced CT acquisition.

\section*{Conflict of Interest}
The authors declare that they have no conflict of interest.

\bibliography{paper_GianmarcoSantini}

\begin{table}[h]
\footnotesize
        \begin{tabular}{ll}
        &Author: Gianmarco Santini\\
        &Institute: University of Pisa\\
        &Street: \\
        &City: Pisa\\
        &Country:  Italy\\
        &Email: gianmarco.santini@ing.unipi.it\\
        \end{tabular}
\end{table}

\end{document}